\pdfoutput=1

\documentclass[11pt]{article}

\usepackage[preprint]{acl}
\usepackage{multicol}
\usepackage{multirow}
\usepackage{times}
\usepackage{latexsym}
\usepackage{booktabs}
\usepackage{amssymb}
\usepackage{bbm}
\usepackage[T1]{fontenc}

\usepackage[utf8]{inputenc}
\usepackage{amsmath}
\usepackage{microtype}

\usepackage{inconsolata}

\usepackage[capitalize,noabbrev]{cleveref}

\usepackage{enumitem}
\usepackage{xcolor}
\usepackage{tcolorbox}
\tcbuselibrary{skins,breakable}
\usepackage{url}
\usepackage{subfigure}
\usepackage{listings}
\usepackage{titlesec}
\usepackage{multirow}
\usepackage{makecell}
\usepackage{cuted}
\usepackage{array}       
\usepackage{caption}     
\usepackage{dsfont}

\usepackage{algorithm}
\usepackage{algpseudocode}
\usepackage{amsmath}
\usepackage{amssymb}

\usepackage{enumitem}
\newenvironment{itemize*}%
 {\leftmargini=20pt\begin{itemize}%
  \setlength{\itemsep}{3pt}%
  \setlength{\parskip}{0pt}%
  }%
 {\end{itemize}} 
\newenvironment{enumerate*}%
 {\begin{enumerate}%
  \setlength{\itemsep}{0pt}%
  \setlength{\parskip}{0pt}}%
 {\end{enumerate}}

\DeclareTextCommand{\textquotedbl}{OT1}{\char`\"}
\lstdefinestyle{jsonTiny}{
  basicstyle=\ttfamily\scriptsize,
  breaklines=true,
  breakindent=0pt,
  columns=fullflexible,
  keepspaces=true,
  showstringspaces=false,
  upquote=true,
  frame=none
}

\definecolor{midnightgreen}{rgb}{0.0, 0.29, 0.33}
\definecolor{deepgreen}{HTML}{0aa344}
\definecolor{deeppurple}{HTML}{7030a0}
\definecolor{deepblue}{HTML}{171d91}
\definecolor{brown}{HTML}{843c0c}
\definecolor{shadered}{HTML}{ffe5e5}
\definecolor{shadegreen}{HTML}{e5f7ed}
\definecolor{msftBlack}{RGB}{0,0,0}
\definecolor{lightred}{RGB}{255,163,163}
\definecolor{deepred}{RGB}{146,0,0}
\definecolor{rqBlueBg}{HTML}{EAF4FF}
\definecolor{tzBlueHeader}{RGB}{78,160,205}
\definecolor{tzBlueHeader2}{RGB}{105,185,225}
\definecolor{tzBlueBorder}{RGB}{115,190,225}
\definecolor{tzBlueFill}{RGB}{232,246,252}
\definecolor{rqBlueBorder}{HTML}{6AADE4}

\newcommand{\green}{\textcolor{deepgreen}}

\usepackage{pifont}

\NewDocumentCommand{\heng}
{ mO{} }{\textcolor{red}{\textsuperscript{\textit{Heng}}\textsf{\textbf{\small[#1]}}}}
\NewDocumentCommand{\cheng}
{ mO{} }{\textcolor{orange}{\textsuperscript{\textit{Cheng}}\textsf{\textbf{\small[#1]}}}}
\NewDocumentCommand{\jy}
{ mO{} }{\textcolor{orange}{\textsuperscript{\textit{Jiayu}}\textsf{\textbf{\small[#1]}}}}

\title{UserHarness: Harnessing User Minds for Stronger Agent Theory-of-Mind}

\author{
\textbf{Cheng Qian},
\textbf{Jiayu Liu},
\textbf{Heng Ji} \vspace{1.5mm}\\
University of Illinois Urbana-Champaign \vspace{1mm}\\
\texttt{\{chengq9,jiayul12,hengji\}@illinois.edu} \vspace{0mm}\\
}

\begin{document}
\maketitle


\begin{figure*}
    \centering
    \vspace{-0mm}
    \includegraphics[width=\linewidth]{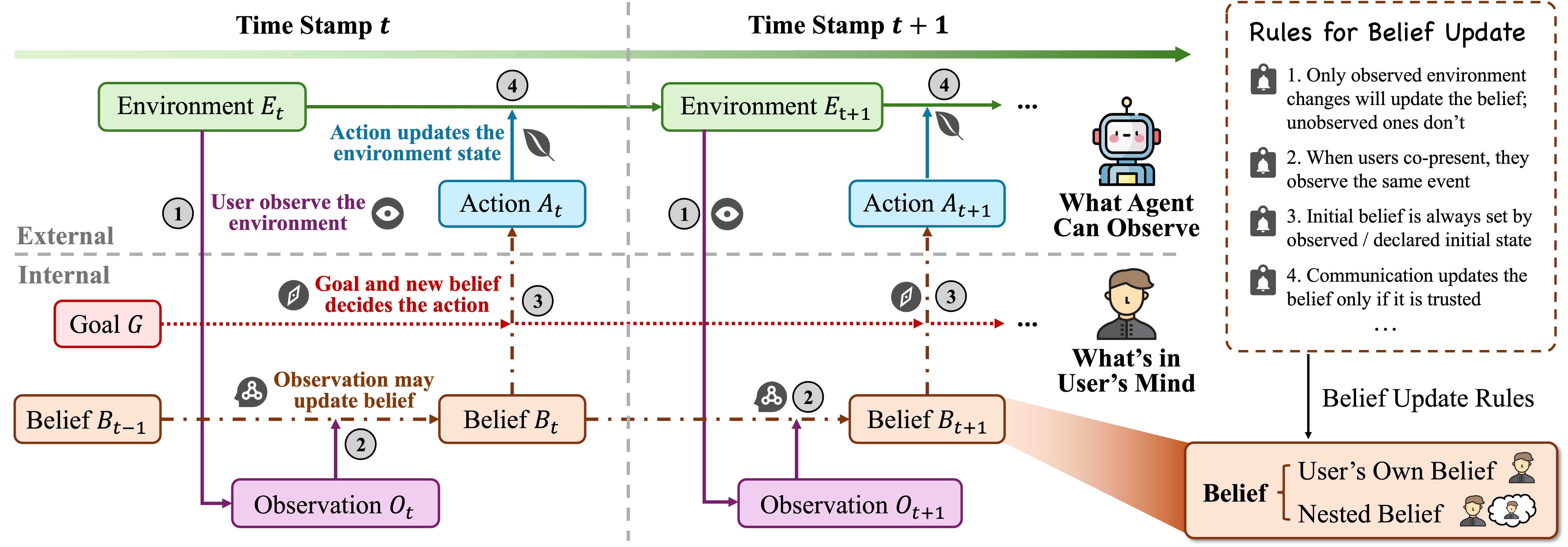}
    \vspace{-0mm}
    \caption{\textbf{UserHarness explicitly models the user mind as a temporally evolving belief--goal--action loop for Theory-of-Mind evaluation}. At each time step, the user observes the external environment $E_t$, forms an observation $O_t$, and updates their internal belief from $B_{t-1}$ to $B_t$. The updated belief, together with the user's goal $G$, determines the user's action $A_t$, which in turn changes the environment to $E_{t+1}$. UserHarness enables the tested agent to explicitly track this hidden cognitive process across time, including both the user's own belief and nested beliefs about other agents' mental states. The locked elements indicate private or inaccessible mental states that must be inferred rather than directly observed. By reconstructing these belief dynamics, the tested agent can better understand the user's perspective and improve performance on ToM questions.}
    \vspace{-0mm}
    \label{fig:method}
\end{figure*}


\newcommand{\env}{\mathcal{E}}
\newcommand{\obs}{\mathcal{O}}
\newcommand{\bel}{\mathcal{B}}
\newcommand{\act}{\mathcal{A}}
\newcommand{\goal}{\mathcal{G}}
\newcommand{\users}{\mathcal{U}}
\newcommand{\agent}{\mathcal{M}}
\newcommand{\todo}[1]{\textcolor{red}{[TODO: #1]}}

\begin{abstract}
Understanding what a user believes and intends is central to building effective agent assistants. This ability is often evaluated through Theory-of-Mind (ToM) tasks, where success requires reasoning from the user’s perspective. However, many existing approaches address ToM with complex pipelines that model behavior indirectly, without explicitly reconstructing the user’s mental state. This misses the core structure of the problem: users act based on their beliefs, which are updated through observations of the environment; beliefs and intentions jointly determine actions, which in turn change the environment; and social reasoning often requires nested beliefs about what others believe or intend.
We propose \textbf{UserHarness}, a simple framework that reframes ToM reasoning as explicit user-mind reconstruction. UserHarness decomposes the user’s mental state, its relation to the external environment, and the actions that follow from it, enabling agents to track what the user observes, believes, intends, and does. Across five benchmarks, UserHarness reaches up to 95.94\% macro accuracy, improving over existing inference methods by more than 15\% relative and over the strongest prompt-only harness by about 20\% relative. These results suggest that robust user understanding requires reasoning from the roots of the user’s mind, positioning user harnessing as a promising foundation for more adaptive future assistants.
\end{abstract}

\section{Introduction}

A truly useful agent assistant must understand more than a user's surface request. It should infer what the user knows, what the user may have missed, what the user intends, and why the user takes certain actions. This ability is often studied through Theory-of-Mind (ToM) tasks, which test whether an agent can reason about beliefs, goals, intentions, and social relations from another person's perspective \citep{Premack1978Does,Wimmer1983Beliefs,Baker2017Rational,Kim2023FANToM}.

The core difficulty of ToM does not come merely from long context or complex language, but often lie in perspective: the user acts from their own mental state, instead of the full state of the world \citep{Wimmer1983Beliefs,Baker2009Action,Baker2017Rational}. A user may search in the wrong place because they hold an outdated belief. A user may follow a misleading message because it changes what they believe. A user may reason about what another person knows, creating nested beliefs that are not visible from the surface story alone \citep{Perner1985John,Kim2023FANToM,Wu2023HITOM,Sclar2023Minding}. To answer such questions correctly, an assistant agent must recover the mental structure that connects environment, observation, belief, intention, and action \citep{Baker2009Action,JaraEttinger2016Naive,Baker2017Rational,Kim2025Hypothesis}.

Existing approaches often aim to improve ToM performance through stronger prompting, better theories, or more elaborate behavioral reasoning pipelines \citep{Wilf2024Think,Jung2024Perceptions,Hou2024TimeToM,Kim2025Hypothesis,zhang2026autotom}. Although useful, these methods often leave the core issue unresolved: they ask the model to produce better answers without requiring it to explicitly reconstruct the mind of the user in question. We argue that ToM reasoning should instead begin with this reconstruction. Before predicting what a user will believe, intend, or do, an agent should infer what the user observes in the environment, how those observations update their beliefs, how beliefs and intentions jointly shape actions, and how those actions in turn alter the environment. In social scenarios, this process must also account for nested beliefs about what other users believe, intend, or know \citep{Perner1985John,Kim2023FANToM,Wu2023HITOM}.

Based on this view, we propose \textbf{UserHarness}, a simple inference-time framework that reframes ToM solving as \textbf{user-mind reconstruction}. Rather than asking the tested agent to answer directly from the raw story, UserHarness guides it to recover the user's perspective: what the user has observed, how those observations shape their beliefs and intentions, and how this mental state explains or predicts action. The framework is grounded in a few core principles: users act on beliefs rather than reality; beliefs update only through accessible evidence; intentions operate under those beliefs; and social reasoning may require tracking how one user's words or actions affect another one's mind \citep{Wimmer1983Beliefs,Baker2009Action,Baker2017Rational,Jung2024Perceptions}. By making this structure explicit, UserHarness separates the true environment from the user's subjective view and turns ToM from direct narrative prediction into a principled process of reconstructing the user's mind in relation to the environment and other users.

We evaluate UserHarness across five representative ToM benchmarks that cover false belief, nested belief, action prediction, goal inference, communication, and social intention. We reveal that UserHarness substantially improves performance over existing baselines. In the main setting, UserHarness reaches 95.94\% macro accuracy, improving over existing model-inference methods and prompt baselines by more than 15\% relative. This gain also generalizes across model families: UserHarness reduces the performance spread across backbones from 26.75 points under direct prompting to 3.65 points, with every evaluated backbone exceeding 92\% macro accuracy. These results suggest that explicit user-mind reconstruction is more effective than relying on increasingly complex prompting or behavioral heuristics. To summarize, our contributions are threefold:
\begin{itemize}[topsep=0pt, partopsep=0pt, leftmargin=*, itemsep=-3pt]
    \item We introduce UserHarness, an inference-time scaffold that converts ToM evaluation from answer prediction into user-mind reconstruction.
    \item We formalize user reasoning as a perception--belief--action loop that separates environment, observation, belief, intention, and action.
    \item We show across benchmarks that UserHarness consistently improves over prompt and inference baselines, with ablations highlighting the value of constrained model participation.
\end{itemize}

\section{Related Works}

\paragraph{Enhancing Language Agent's Theory of Mind.}
Theory-of-Mind research originates in the study of how humans attribute mental states such as beliefs, desires, intentions, perceptions, and knowledge to others\citep{Premack1978Does,Wimmer1983Beliefs,BaronCohen1985Does,Perner1985John,Baker2009Action,Baker2017Rational}. In LLMs, recent evaluations show a mixed picture: models can solve some standard false-belief tasks, but their behavior remains brittle under task perturbations, adversarial controls, higher-order beliefs, interactive dialogue, and grounded action choices \citep{Sap2022Neural,Kosinski2024Evaluating,Ullman2023Large,Shapira2024Clever,Le2019Revisiting,Ma2023Towards,Kim2023FANToM,Wu2023HITOM,Zhou2023How}. To improve ToM reasoning, recent work has introduced perspective-taking prompts such as SimToM \citep{Wilf2024Think}, perception-to-belief scaffolding such as PercepToM \citep{Jung2024Perceptions}, temporal belief-state construction such as TimeToM \citep{Hou2024TimeToM}, explicit symbolic belief trackers such as SymbolicToM \citep{Sclar2023Minding}, and hypothesis-tracking methods such as ThoughtTracing \citep{Kim2025Hypothesis}; related approaches also use multi-agent scaffolding and adversarial data generation to expose or improve social reasoning \citep{Cross2025Hypothetical,Sclar2024Explore}. UserHarness is positioned differently: it treats ToM solving as explicit user-mind reconstruction that links observation, belief, intention, action, and nested belief rather than as a general prompting.

\paragraph{User-Centric Agent Training and Benchmarking.}
User-centric agent research has long emphasized that helpful systems should infer user goals, plans, preferences, and information needs rather than merely execute literal commands \citep{Kautz1986Generalized,Horvitz1999Principles,Kobsa2001Generic,Amershi2019Guidelines}. Recent LLM-agent benchmarks have moved from static instruction following toward interactive environments such as web browsing, operating-system control, workplace software, app ecosystems, and tool-use sandboxes \citep{Zhou2023WebArena,Drouin2024WorkArena,Xie2024OSWorld,Qian2025Interactive,Lu2025ToolSandbox}; at the same time, user-facing benchmarks increasingly evaluate whether agents can satisfy real user intents, converse with simulated users, follow domain policies, discover latent preferences, and remain reliable across multi-turn trajectories \citep{Wang2024User,qian2024tell,Yao2024Tau,qian2025userrl}. These efforts make agent evaluation more realistic, but they usually focus on task completion, policy compliance, tool state, or preference satisfaction rather than on reconstructing the user's subjective mental state. UserHarness complements this line by making user-mind reconstruction the central reasoning object for ToM-style agent assistance.

\section{Method}
Theory-of-Mind (ToM) evaluation asks a model to answer from the perspective of a situated user. However, the tested model is usually given the full story, while the user in the story only has partial access to events. This mismatch often causes \emph{perspective leakage}: the model answers using the true world state instead of the user's belief.

We propose \textbf{UserHarness}, a pure inference-time framework that constrains the tested agent to reason through the user's mental state before answering. Rather than treating a ToM problem as direct prediction from story to answer, UserHarness turns it into a structured inference problem: recover what the user could observe, update what the user should believe, infer what action or intention follows from that belief, and then verify which answer option is consistent with this trace.

The key idea is that ToM questions should be solved from the user's epistemic position, not from the model's privileged view of the narrative. UserHarness provides this epistemic scaffold without training or modifying the tested model.

\subsection{User-Mind Harnessing}

Let $\mathcal{U}$ be the set of users in a scenario. At time $t$, the external state is denoted by $E_t$, which may contain locations, objects, visible events, and communication events. A user $u \in \mathcal{U}$ does not observe $E_t$ directly. Instead, the user receives only part of  $E_t$ as partial observation

\begin{small} \begin{equation*}
O_t^u = \Omega(E_t,u),
\end{equation*} \end{small}
where $\Omega$ describes what is accessible to $u$ at that moment, such as events in the same room, visible object changes, or messages addressed to the user.

UserHarness maintains a belief state $B_t^u$ for each user. This state contains the user's own belief and, when required, nested beliefs about other users:

\begin{small} \begin{equation*}
B_t^u = \bigl(b_t^u,\; b_t^{u\rightarrow v},\; b_t^{u\rightarrow v\rightarrow w},\ldots\bigr).
\end{equation*} \end{small}
Here, $b_t^u$ denotes what $u$ believes about the world, while $b_t^{u\rightarrow v}$ denotes what $u$ believes $v$ believes. The nesting depth is dynamically determined by the benchmark question.

User's belief may change with observation, and this update is governed by a rule-guided operator:

\begin{small} \begin{equation*}
B_t^u = \Gamma_{\mathcal{R}}(B_{t-1}^u, O_t^u).
\end{equation*} \end{small}
The rule set $\mathcal{R}$ enforces perspective constraints. As shown in \Cref{fig:method}, it encodes simple but crucial rules such as only observed changes update a user's beliefs, accessible communication updates only its recipients, etc. Thus, user beliefs are updated from accessible evidence rather than the hidden true state, preventing privileged story information from leaking into the modeled user mind.

Given the updated belief and the user's goal $G^u$, UserHarness models the user's action as

\begin{small} \begin{equation*}
A_t^u = \pi(G^u, B_t^u).
\end{equation*} \end{small}
The action then changes the external state:

\begin{small} \begin{equation*}
E_{t+1}=T(E_t,A_t^u).
\end{equation*} \end{small}
Therefore, the framework follows the loop

\begin{small} \begin{equation*}
E_t \rightarrow O_t^u \rightarrow B_t^u \rightarrow A_t^u \rightarrow E_{t+1},
\end{equation*} \end{small}
which is repeated until the benchmark question is reached.
At the final step, the question is answered by querying the constructed trace

\begin{small} \begin{equation*}
\tau^u = \{(E_t,O_t^u,B_t^u,A_t^u)\}_{t=1}^{T}.
\end{equation*} \end{small}
For multiple-choice tasks, each option is treated as a candidate claim about the user's belief, action, goal, or intention. UserHarness selects the option that is consistent with $\tau^u$ and rejects options that require the user to know unobserved facts, hold unsupported beliefs, or act against their current belief and goal.

This formulation makes the role of the harness explicit. The tested agent is not asked to answer directly from the full narrative. Instead, it is guided to reconstruct the user's mental trajectory and answer from that trajectory. The result is an inference-time scaffold that aligns the model's reasoning process with the structure of ToM evaluation.

\begin{algorithm}[!t]
\small
\caption{\vspace{-0.8mm}\textsc{UserHarness}: Core Steps of User Mind Reconstruction}
\label{algo:user_mind_reconstruction}
\begin{algorithmic}[1]
\Require Story events $\{e_t\}_{t=1}^{T}$, users $\mathcal{U}$, target user $u^\star$
\Require Initial environment $E_1$, user goals $\{G^u\}_{u\in\mathcal{U}}$, perspective rules $\mathcal{R}$
\Require Observation function $\Omega$, belief updater $\Gamma_{\mathcal{R}}$, policy $\pi$, transition function $T$

\ForAll{$u\in\mathcal{U}$}
    \State Initialize user mind $B_0^u=(b_0^u,\{b_0^{u\rightarrow v\rightarrow\cdots}\})$
    \State Initialize trace $\tau^u\gets\emptyset$
\EndFor

\For{$t=1,\dots,T$}
    \ForAll{$u\in\mathcal{U}$}

        \Statex \textbf{Step 1: Observe environment}
        \State $O_t^u \gets \Omega(E_t,u;\mathcal{R})$
        \Comment{Extract only what user $u$ can observe}

        \Statex \textbf{Step 2: Update user mind}
        \State $B_t^u \gets \Gamma_{\mathcal{R}}(B_{t-1}^u,O_t^u)$
        \Comment{Update own belief and nested beliefs}

        \Statex \textbf{Step 3: Decide action from goal and belief}
        \State $A_t^u \gets \pi(G^u,B_t^u)$
        \Comment{Act according to the user's belief, not the true state}

        \State $\tau^u \gets \tau^u \cup \{(E_t,O_t^u,B_t^u,A_t^u)\}$

    \EndFor

    \Statex \textbf{Step 4: Update environment}
    \State $E_{t+1}\gets T(E_t,\{A_t^u\}_{u\in\mathcal{U}},e_t)$
    \Comment{Actions and story events change the world state}

\EndFor

\State \Return Target user mind trace $\tau^{u^\star}$
\end{algorithmic}
\end{algorithm}

\subsection{Inference-Time Proving}

Given a benchmark instance $x=(s,q,\mathcal{Y})$, where $s$ is the story, $q$ is the question, and $\mathcal{Y}$ is the answer set, UserHarness first parses the story into event steps and builds the trace $\tau$. The final question is then mapped to a query over the trace:

\begin{small}
\begin{equation*}
    \phi_q(\tau) \in \{\text{belief}, \text{nested belief}, \text{goal}, \text{action}, \text{intention}\}.
\end{equation*}
\end{small}
For multiple-choice tasks, each option $y_i \in \mathcal{Y}$ is treated as a candidate claim. UserHarness selects the answer by checking consistency:

\begin{small} \begin{equation*}
\hat{y} = \arg\max_{y_i \in \mathcal{Y}} \mathrm{Consistent}(y_i, \tau, q).
\end{equation*} \end{small}
An option is rejected if it contradicts the user's observation history, belief state, communication access, goal, or action rule. This converts ToM answering into an option-level proof problem grounded in the user's mental trajectory.

Instead of directly reasoning to an answer, the tested agent now operates within the scaffold: it helps interpret the story, identify relevant users and events, audit candidate traces, and handle incomplete symbolic proofs. Because the process is anchored to the trace $\tau$, the final decision remains tied to the user's perspective. Later experiments show that this lowers the model's reasoning burden and significantly improves performance.

\paragraph{Why UserHarness Works.}
UserHarness improves ToM reasoning by separating three quantities that direct prompting often conflates: the true environment $E_t$, the user's belief $B_t^u$, and nested beliefs $B_t^{u \rightarrow v \rightarrow \cdots}$. This separation is crucial because many ToM benchmarks are designed precisely around cases where these states diverge.

In false-belief problems, the correct answer depends on an outdated belief rather than the true object location. In communication problems, only agents who hear or accept a message should update their beliefs. In nested-belief problems, the answer depends not only on what happened, but on who observed whom observing it. UserHarness makes these distinctions explicit and forces the tested agent to reason through them before selecting an answer.

The framework therefore improves performance but by changing the structure of inference: the tested agent must justify the answer through the user's accessible mental state. For the instantiation of UserHarness on different benchmarks, please refer to \Cref{apdx:benchmark_instantiations} for details.

\section{Experiments}

\subsection{Experiment Settings}
\paragraph{Benchmarks and models.}
We evaluate the effectiveness of UserHarness on five widely used ToM benchmarks: BigToM~\citep{Gandhi2023Understanding}, Hi-ToM~\citep{Wu2023HITOM}, ToMi~\citep{Le2019Revisiting}, MMToM-QA~\citep{jin2024mmtom}, and MuMA-ToM~\citep{shi2025muma}. 
For all benchmarks, we use the text-only setting to evaluate language agents. These benchmarks cover a broad spectrum of ToM reasoning, ranging from first-order false-belief tracking and belief--reality distinction, to higher-order nested belief reasoning, goal and action prediction, communication-based belief update, and multi-agent social intention inference. 
This diversity allows us to test whether UserHarness provides a general user-mind reconstruction framework rather than a benchmark-specific solver. 
We evaluate UserHarness with diverse models from GPT, Gemini, Claude, Llama, and Qwen families.

\paragraph{Baselines.}
We evaluate our method against a set of baselines that cover three complementary ways of improving Theory-of-Mind reasoning. First, we include prompting-only baselines for each underlying model: \emph{direct prompting}, where the model reasons and answers directly; \emph{paradox prompting}, where the model tests each answer option by assuming it is true and checking for contradictions; and \emph{ledger prompting}, where the model is asked to organize its reasoning around the agent's environment, observations, beliefs, goals, and actions. The ledger baseline provides a stronger prompting control than direct reasoning, but unlike our method it does not impose externally maintained symbolic traces or structured state updates.

Second, we compare against ToM-specific inference scaffolds, including SymbolicToM~\citep{Sclar2023Minding} and SimToM~\citep{Wilf2024Think}. These methods are designed specifically to improve belief reasoning: SymbolicToM augments language models with explicit multi-character belief tracking, while SimToM uses perspective-taking to restrict reasoning to information available to the target agent. Finally, we compare with model-based mental-state inference methods, including BIP-ALM~\citep{jin2024mmtom}, LIMP~\citep{shi2025muma} and AutoToM~\citep{zhang2026autotom}, which infer agents' latent mental states through inverse-planning-style procedures. All decoding is performed with temperature set to zero to ensure deterministic outputs. Please see \Cref{apdx:exp_settings} for more details.

\subsection{Main Results}

\begin{table*}[!t]
\centering
\footnotesize
\setlength{\tabcolsep}{8pt}
\renewcommand{\arraystretch}{1.02}
\resizebox{\linewidth}{!}{
\begin{tabular}{@{}llcccccc@{}}
\toprule
\textbf{Approach} & \textbf{Backbone / Setting}
& \textbf{ToMi} & \textbf{BigToM} & \textbf{MMToM-QA}
& \textbf{MuMA-ToM} & \textbf{Hi-ToM} & \textbf{Macro} \\
\midrule
\addlinespace[0.35em]
\multicolumn{8}{@{}l}{\textit{Prompting controls on our evaluated backbones}} \\
\midrule
\multirow{8}{*}{Direct prompting}
& Qwen3-8B       & 54.52 & 72.33 & 37.17 & 44.56 & 50.92 & 51.90 \\
& Qwen3-14B      &  6.20 & 73.50 & 37.17 & 56.44 & 56.00 & 45.86 \\
& Qwen3-32B      & 14.02 & 88.50 & 37.50 & 60.44 & 53.42 & 50.78 \\
& Llama-3.1-8B   & 52.48 & 74.42 & 43.33 & 42.33 & 41.08 & 50.73 \\
& GPT-OSS-20B    & 70.08 & 86.75 & 47.33 & 69.67 & 66.83 & 68.13 \\
& GPT-OSS-120B   & 80.22 & 92.00 & 64.17 & 80.78 & 74.92 & 78.42 \\
& GPT-5.4-mini   & 78.36 & 76.42 & 35.00 & 46.33 & 57.08 & 58.64 \\
& GPT-5.4        & 79.72 & 86.92 & 48.67 & 75.67 & 72.08 & 72.61 \\
\midrule
\multirow{8}{*}{Paradox prompting}
& Qwen3-8B       & 17.32 & 76.42 & 39.67 & 53.00 &  0.00 & 37.28 \\
& Qwen3-14B      & 48.96 & 77.08 & 45.17 & 56.44 & 12.67 & 48.06 \\
& Qwen3-32B      & 50.38 & 87.33 & 48.00 & 53.33 &  3.58 & 48.53 \\
& Llama-3.1-8B   &  5.50 & 74.00 & 47.00 & 43.44 &  1.50 & 34.29 \\
& GPT-OSS-20B    & 69.56 & 87.83 & 46.67 & 71.78 & 65.92 & 68.35 \\
& GPT-OSS-120B   & 75.05 & 91.83 & 67.50 & 80.11 & 73.42 & 77.58 \\
& GPT-5.4-mini   & 83.97 & 77.92 & 43.50 & 56.00 & 60.58 & 64.39 \\
& GPT-5.4        & 89.10 & 93.83 & 63.33 & 68.44 & 75.50 & 78.04 \\
\midrule
\multirow{8}{*}{Ledger prompting}
& Qwen3-8B       & 26.44 & 93.50 & 45.67 & 13.56 &  9.75 & 37.78 \\
& Qwen3-14B      & 25.28 & 97.25 & 44.33 & 45.89 & 36.00 & 49.75 \\
& Qwen3-32B      & 47.34 & 96.50 & 47.00 & 36.56 & 24.75 & 50.43 \\
& Llama-3.1-8B   & 11.24 & 92.17 & 47.33 & 29.00 &  8.17 & 37.58 \\
& GPT-OSS-20B    & 62.37 & 96.17 & 48.83 & 67.89 & 67.50 & 68.55 \\
& GPT-OSS-120B   & 77.38 & 97.08 & 69.83 & 78.89 & 75.08 & 79.65 \\
& GPT-5.4-mini   & 73.03 & 96.67 & 34.00 & 49.44 & 58.92 & 62.41 \\
& GPT-5.4        & 85.72 & 98.25 & 62.83 & 64.78 & 77.00 & 77.72 \\
\midrule
\multicolumn{8}{@{}l}{\textit{Existing ToM-specific scaffolds}} \\
\midrule
SymbolicToM & reported setting & 98.60 & \multicolumn{1}{c}{--} & \multicolumn{1}{c}{--} & \multicolumn{1}{c}{--} & 44.50 & \multicolumn{1}{c}{--} \\
SimToM      & reported setting & 79.90 & 77.50 & 51.00 & 47.63 & 71.00 & 65.41 \\
\midrule
\addlinespace[0.35em]
\multicolumn{8}{@{}l}{\textit{Existing model-based mental-state inference}} \\
\midrule
BIP-ALM & reported setting & 55.60 & 50.33 & 56.17 & 33.90 & 14.50 & 42.10 \\
LIMP    & reported setting & 44.60 & 61.67 & 55.33 & 76.60 &  6.50 & 48.94 \\
AutoToM & GPT-4o           & 88.30 & 86.92 & 83.00 & 81.44 & 72.50 & 82.43 \\
\midrule
\multicolumn{8}{@{}l}{\textit{Ours main method}} \\
\midrule
\multirow{6}{*}{\textbf{\makecell[l]{UserHarness\\(Open-source Models)}}}
& Qwen3-8B               & \textbf{100.00} & 89.67 & 97.67 & 95.78 & 86.58 & 93.94 \\
& Qwen3-14B              & \textbf{100.00} & 94.17 & 98.17 & 95.67 & 87.00 & 95.00 \\
& Qwen3-32B              & \textbf{100.00} & 95.42 & 98.00 & 95.78 & \textbf{87.08} & 95.26 \\
& Llama-3.1-8B           & \textbf{100.00} & 82.83 & 96.67 & 95.56 & 86.42 & 92.29 \\
& GPT-OSS-20B            & \textbf{100.00} & 90.17 & 97.00 & 95.67 & 86.83 & 93.93 \\
& GPT-OSS-120B           & \textbf{100.00} & 93.83 & 97.33 & 95.78 & 86.75 & 94.74 \\
\midrule
\multirow{6}{*}{\textbf{\makecell[l]{UserHarness\\(Closed-source Models)}}}
& GPT-5.4-mini           & \textbf{100.00} & 95.25 & 98.17 & 95.67 & 86.58 & 95.13 \\
& GPT-5.4                & \textbf{100.00} & 97.67 & 97.83 & \textbf{95.89} & 86.83 & 95.64 \\
& Gemini-3-Flash         & \textbf{100.00} & 86.25 & 97.50 & 94.89 & 86.42 & 93.01 \\
& Gemini-3.1-Pro         & \textbf{100.00} & 98.25 & 97.83 & \textbf{95.89} & 86.42 & 95.68 \\
& Claude-Sonnet-4.6      & \textbf{100.00} & 98.17 & \textbf{98.50} & 95.78 & 86.25 & 95.74 \\
& Claude-Opus-4.7        & \textbf{100.00} & \textbf{98.67} & 98.33 & \textbf{95.89} & 86.83 & \textbf{95.94} \\
\bottomrule
\end{tabular}
}
\caption{
\textbf{Main evaluation results} across five benchmarks. UserHarness rows report our main setting, with different open- and closed-source model participation in harness translation, proof, and audit with constraints. Dashes indicate that a method is not applicable or reported. Best UserHarness results are shown in \textbf{bold}.
}
\label{tab:main}
\end{table*}


\paragraph{Prompt-only harnesses are not reliable enough.}
As shown in \Cref{tab:main}, stronger prompting alone does not consistently solve the benchmarks. Direct prompting ranges from 45.86\% to 72.61\% macro accuracy across the backbones, with GPT-5.4 clearly outperforming the open-source models but still far below the UserHarness main setting. Paradox and ledger prompting help on some slices, especially BigToM, but often fail on ToMi, Hi-ToM, or MuMA-ToM, where the model must maintain a complete belief state in free-form text. Even GPT-5.4 reaches only 78.04\% with paradox prompting and 77.72\% with ledger prompting, suggesting that prompt wording alone does not provide a stable representation of user-perspective state.

\paragraph{UserHarness substantially improves ToM accuracy.}
UserHarness converts each story into an explicit symbolic representation, proves option-level claims from that representation, and audits the proof for consistency, as illustrated in \Cref{fig:method}. This structured decomposition raises macro accuracy to 95.94\% with Claude Opus, improving over AutoToM by 13.51 macro points. Aggregating the best per-benchmark result across all evaluated backbones yields near-perfect performance, including 100\% on ToMi. These gains suggest that the key benefit is not merely better reasoning effort, but a more reliable externalization of agents' beliefs, observations, and user-specific perspectives.

\paragraph{The improvement generalizes across backbones.}
The gains are not limited to the strongest model. Relative to direct prompting, UserHarness improves macro accuracy by 49.14 points for Qwen3-14B and still adds 23.03 points for GPT-5.4, the best direct-prompt solver. The larger gains for smaller models indicate that the harness compensates for limited free-form reasoning capacity, while the remaining gain for GPT-5.4 shows that even strong models benefit from explicit state tracking. Thus, UserHarness improves both weak and strong backbones by shifting the burden from implicit narrative reasoning to structured verification.

\paragraph{UserHarness narrows, but does not erase, the model-size gap.}
Under direct prompting, macro accuracy varies by 26.75 points across the backbones for which we have prompt-only runs, from 45.86\% for Qwen3-14B to 72.61\% for GPT-5.4. With UserHarness, this spread shrinks to only 3.65 points across all eight evaluated backbones, from 92.29\% for Llama-3.1-8B to 95.94\% for Claude Opus, while every backbone exceeds 92\% macro accuracy. This compression shows that the scaffold reduces the cognitive load of open-ended ToM solving by breaking it into translation, proof, and audit steps. At the same time, the residual spread indicates that the tested model's own translation and verification capabilities still matter.

\section{Analysis and Ablations}

\paragraph{Symbolic Harness vs. Model Harness.}

\begin{table}[!t]
  \centering
  \footnotesize
  \setlength{\tabcolsep}{5pt}
  \resizebox{\linewidth}{!}{
  \begin{tabular}{@{}lccc@{}}
  \toprule
  \textbf{Benchmark} & \textbf{Symbolic UserHarness} & \textbf{Model UserHarness} & \textbf{Gap} \\
  \midrule
  ToMi & 100.00 & 100.00 & 0.00 \\
  BigToM & 6.75 & 95.42 & \green{$\uparrow$ 88.67} \\
  MMToM-QA & 91.33 & 98.00 & \green{$\uparrow$ 6.67} \\
  MuMA-ToM & 93.67 & 95.78 & \green{$\uparrow$ 2.11} \\
  Hi-ToM & 86.42 & 87.08 & \green{$\uparrow$ 0.66} \\
  \midrule
  Macro & 75.63 & 95.26 & \green{$\uparrow$ 19.63} \\
  \bottomrule
  \end{tabular}
  }
  \caption{\textbf{Pure symbolic versus model participated UserHarness.} Model participated UserHarness uses Qwen3-32B as the analysis anchor for comparison.}
  \vspace{-2mm}
  \label{tab:symbolic_gap}
\end{table}

Since the model performances in \Cref{tab:main} under UserHarness are close to one another, we ask a more diagnostic question: how much of the improvement comes from the harness scaffold itself, and how much still depends on the model's own capabilities? To answer this, we construct a pure harness-only pipeline that excludes model participation entirely. This pipeline follows the structure of \Cref{fig:method}, but performs translation, parsing, and ToM proof generation through rule-based procedures. The harness is refined using the test inputs, but not the answers, so it may exploit dataset regularities through heuristics but does not access labels. In this setting, the symbolic harness answers only through rule-derived proofs, abstentions, or default behaviors, using the same benchmark test sets.

Using Qwen3-32B as the comparison anchor, \Cref{tab:symbolic_gap} shows that the contribution of model capability is highly benchmark-dependent. On ToMi, the symbolic harness already achieves perfect accuracy, suggesting that a carefully designed rule-based scaffold under UserHarness is sufficient for this relatively simple benchmark. In contrast, BigToM shows a large gap between the symbolic and model paticipation, indicating that model-driven translation, auditing, and error correction remain crucial for more complex settings. These results support three observations: (1) model capability still affects UserHarness performance, although its importance varies substantially across benchmarks; (2) the strong performance of the symbolic harness on several datasets shows that the harness structure itself can substantially reduce dependence on raw model ability. (3) these findings raise a broader concern: some existing ToM benchmarks may be vulnerable to deliberate symbolic scaffolding, and therefore may overestimate genuine model-level theory-of-mind competence.

\paragraph{Audit Override Calibration.}

\begin{figure}[t]
  \centering
  \includegraphics[width=\linewidth]{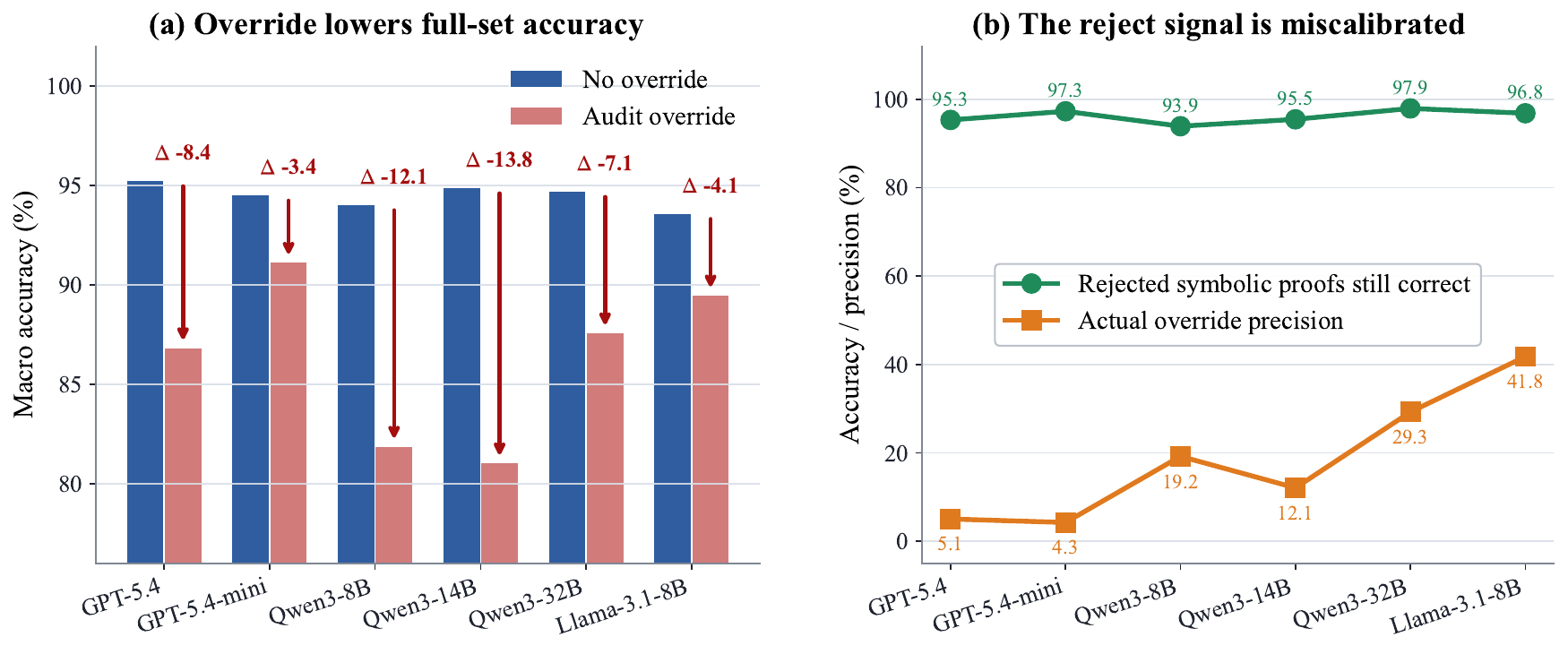}
  \caption{\textbf{Audit calibration across models.} 
  \textbf{Left:} Allowing models to freely override UserHarness proofs consistently reduces macro accuracy across backbones. 
  \textbf{Right:} This degradation reflects poor audit calibration, as many proofs rejected by the model are in fact correct.}
  \label{fig:calibration}
\end{figure}

We next examine whether models can reliably audit the ToM reasoning proofs without any harness. In the ablation, the tested model is allowed to freely replace the harness-derived proof and final answer whenever it judges the reasoning trace to be flawed. As shown in \Cref{fig:calibration}, this consistently lowers macro accuracy across backbones. The result suggests that models often react to local uncertainty or plausible alternative interpretations, even when the completed UserHarness proof is correct under the explicit observation and belief trace.

The right panel of \Cref{fig:calibration} explains this failure more directly. We measure two quantities: the fraction of model-rejected proofs that are actually correct, and the fraction of truly incorrect proofs that the model successfully detects and overrides. The former is high, while the latter is low. This indicates that: (1) unconstrained model auditing is poorly calibrated and should not be trusted without an explicit harness, and (2) stronger backbones do not necessarily yield better audit reliability, since they may still fail to detect genuinely incorrect proofs while over-trusting their own reasoning.

\paragraph{Effectiveness of Compute.}

\begin{figure}[!t]
  \centering
  \includegraphics[width=\linewidth]{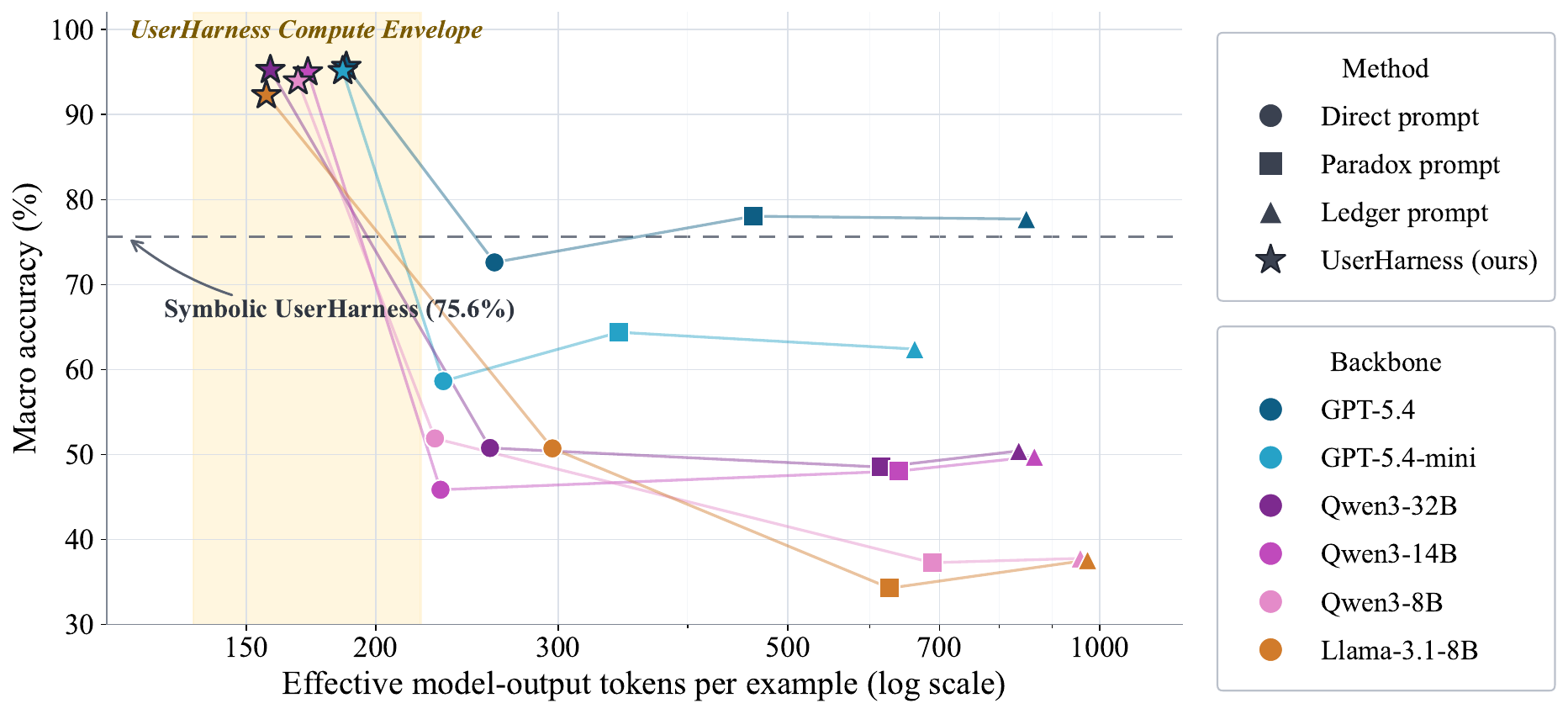}
  \caption{\textbf{Accuracy--compute tradeoff.} UserHarness achieves high macro accuracy within a narrow low-compute envelope, while prompt-only methods often spend more effective output tokens with substantially lower accuracy.}
  \label{fig:compute}
\end{figure}

\begin{figure*}[!t]
  \centering
  \includegraphics[width=\linewidth]{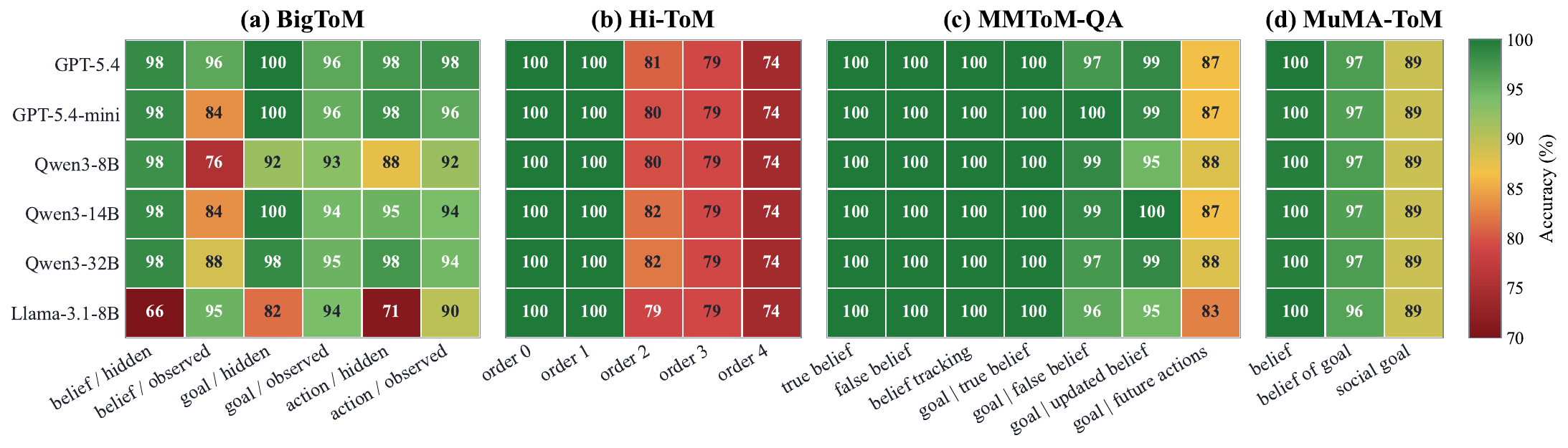}
  \caption{\textbf{Performance across difficulty tiers.} UserHarness achieves consistently strong performance on many explicit belief, goal, and low-order reasoning cases, but residual errors concentrate in structurally harder regimes including higher-order belief, future-action prediction, social-goal inference, etc.}
  \label{fig:difficulty_tiers}
\end{figure*}

We examine whether UserHarness gains come from spending more model-output compute. \Cref{fig:compute} shows that UserHarness occupies a compact and favorable region: across all six backbones, it uses only about 156--187 effective output tokens per example while reaching 92.3--95.6\% macro accuracy. By contrast, prompt-only methods often use many more tokens but achieve substantially lower and more variable performance. This suggests that longer free-form reasoning does not reliably improve ToM accuracy; in several cases, more verbose ledger-style prompting increases compute while leaving accuracy unchanged or worse. The symbolic UserHarness baseline further clarifies the source of the gain: rules alone already reach 75.6\%, but constrained model participation raises performance far beyond this level without giving the model unrestricted control. These results show that: (1) UserHarness improves accuracy by limiting model authority and routing reasoning through a protected symbolic scaffold, rather than by eliciting longer outputs; and (2) future gains are more likely to come from improving the harness interface including translation, verification, and symbolic representation rather than from simply spending more tokens.

\paragraph{Difficulty-Tier and Error Analysis.}

We further analyze where UserHarness succeeds and where it still fails by breaking down performance across benchmark-specific difficulty tiers. As shown in \Cref{fig:difficulty_tiers}, UserHarness is highly effective on many core ToM categories. Across models, it reaches near-perfect accuracy on explicit belief and goal reasoning, and it also perfectly solves low-order belief reasoning. This suggests that once observations and belief states are converted into an explicit trace, UserHarness can reliably execute many forms of structured ToM inference. The gains are also relatively stable across backbones, indicating that the harness does not merely amplify the strongest models, but provides a scaffold that weaker and stronger models can both exploit.

At the same time, the remaining errors concentrate in tiers that require more complex belief-state tracking. In Hi-ToM, performance drops sharply for higher-order belief nesting, showing that recursive mental-state attribution remains difficult even with explicit harness support. In MMToM-QA, future-action prediction remains noticeably weaker, suggesting that translating beliefs into downstream behavior introduces an additional planning or pragmatics bottleneck. In MuMA-ToM, the persistent gap on social-goal inference indicates that socially grounded intent may not be fully captured by UserHarness belief traces alone. These patterns suggest that: (1) UserHarness is effective at stabilizing explicit ToM reasoning and reducing model-dependent variance; and (2) future improvements should focus on richer representations for recursive beliefs, action prediction, and socially grounded goals, where current harness proofs still leave ambiguity unresolved.

\section{Conclusion}
UserHarness reframes ToM reasoning as explicit user-mind reconstruction, shifting the tested agent from answer generation to structured reasoning over observations, beliefs, goals, actions, and nested mental states. Across diverse ToM benchmarks, this perspective-centered scaffold consistently improves reasoning reliability, showing that robust user understanding depends on separating what is true in the world from what the user can observe, believe, and intend. Looking forward, we view UserHarness as a stepping stone toward richer open-domain user-centric interaction, where assistants must move beyond solving controlled ToM tasks to continuously tracking users’ evolving beliefs, inferring socially grounded intentions, and reasoning about recursive mental states under real-world ambiguity.

\bibliography{custom}


\clearpage
\newpage
\appendix

\section*{Appendix}
\label{sec:appendix}

\section{Benchmark Instantiations}
\label{apdx:benchmark_instantiations}

In this section, we describe how each benchmark is instantiated inside the UserHarness loop in \Cref{fig:method}. The goal is to make explicit how each dataset is viewed through the same variables: external environment $E_t$, user observation $O_t$, belief $B_t$, goal $G$, action $A_t$, and the next environment state $E_{t+1}$. Across all benchmarks, answer labels are used only for scoring after prediction. Multiple-choice options are treated as candidate claims about the reconstructed trace, and the selected answer is the option that is most consistent with the user's accessible mental state.

\paragraph{Shared instantiation.}
For every example, UserHarness constructs a transition trace of the form
\[
E_t \rightarrow O_t^u \rightarrow B_t^u \rightarrow A_t^u \rightarrow E_{t+1}.
\]
The environment $E_t$ contains the objective story state: object locations, rooms, containers, user positions, utterances, causal changes, and task-relevant preconditions. The observation $O_t^u$ contains only information accessible to user $u$, such as events in the same room, visible object movements, or messages addressed to the user. The belief $B_t^u$ is updated from observation, not from hidden truth. When the benchmark asks what one user thinks another user thinks, $B_t^u$ is expanded into nested beliefs $B_t^{u\rightarrow v\rightarrow \cdots}$. The goal $G^u$ is either explicitly stated in the story, implied by the question type, or inferred from the user's task trajectory. The action $A_t^u$ is then interpreted as the behavior that follows from $G^u$ and $B_t^u$, not necessarily from the real environment.

The same belief-update principles are used throughout:
\begin{itemize}[topsep=0pt, partopsep=0pt, leftmargin=*, itemsep=-3pt]
    \item \textbf{Observed change updates belief.} If a user observes an object movement, state change, visible outcome, or relevant utterance, the user's belief is updated to reflect that accessible evidence.
    \item \textbf{Unobserved change preserves belief.} If a user is absent, occluded, or not addressed by a communication event, the user's prior belief is carried forward.
    \item \textbf{Co-observation supports nested belief.} If multiple users jointly observe an event, each can believe that the others also observed it. Nested updates stop once a user leaves the shared observation context.
    \item \textbf{Communication is scoped by access.} A message updates only users who hear or receive it, and public/private communication determines which nested belief paths can be updated.
    \item \textbf{Action follows belief and goal.} Search, use, help, hinder, or proceed actions are derived from what the user believes and wants, even when that belief differs from the true environment.
    \item \textbf{Distractors do not update target beliefs.} Events about unrelated objects, rooms, or users are ignored unless they change the queried user's observation, belief, goal, or action precondition.
\end{itemize}

\paragraph{Tested-agent role in the main setting.}
In our main setting, the tested model participates in every example. Its role switches from directly answering the question to help instantiate the ToM claim validation loop. First, the agent is used as a translation helper: it identifies the relevant users, objects, observations, communications, goals, actions, and option-level claims. Second, when the harness has a candidate proof, the agent audits whether the proof is consistent with the environment--observation--belief trace. If the harness cannot produce a symbolic proof, the agent is used as a fallback solver under the same perspective constraints. The final decision remains grounded in the trace unless the symbolic branch abstains.

\begin{tcolorbox}[
  enhanced,
  breakable,
  width=0.98\linewidth,
  colback=white,
  colframe=brown,
  boxrule=1.2pt,
  arc=6pt,
  left=5pt,right=5pt,top=4pt,bottom=2pt,
  title={\small Core translation-helper instruction excerpt},
  coltitle=white,
  colbacktitle=brown,
  fonttitle=\bfseries,
]
\small
\begin{lstlisting}[style=jsonTiny]
Translate the story into prover-readable facts, not an intuitive guess. Extract agents, target objects, relevant observations, environment changes, communications, beliefs, goals, actions, and the option text that matches the extracted state. Apply the rules strictly: observed changes update beliefs; unobserved changes do not; agents act from beliefs and goals, not hidden reality; communication affects only listeners who hear it.
\end{lstlisting}
\end{tcolorbox}

\begin{tcolorbox}[
  enhanced,
  breakable,
  width=0.98\linewidth,
  colback=white,
  colframe=brown,
  boxrule=1.2pt,
  arc=6pt,
  left=5pt,right=5pt,top=4pt,bottom=2pt,
  title={\small Core proof-auditor instruction excerpt},
  coltitle=white,
  colbacktitle=brown,
  fonttitle=\bfseries,
]
\small
\begin{lstlisting}[style=jsonTiny]
Build a transition ledger: environment state, user observations, user beliefs, communications, goals, and actions. For each option, first assume the option is true, then check whether that assumption creates a contradiction with the ledger. Accept a symbolic candidate unless there is a concrete contradiction and another option has no contradiction.
\end{lstlisting}
\end{tcolorbox}

\subsection{BigToM}

BigToM instantiate the single-user realization of the loop. The environment $E_t$ contains the initial state of a target object or condition, an environment change, and often distractor events that should not affect the target user's belief. The observation $O_t^u$ records whether the queried user sees the change itself or a visible consequence of the change. The belief $B_t^u$ is mostly the user's own belief; nested belief is not central in this benchmark. The goal $G^u$ is the task implicit in the question, such as locating, using, choosing, or acting on an item. The action $A_t^u$ is predicted from this goal and the user's current belief.

For belief questions, UserHarness answers from $B_t^u$: an observed change replaces the prior belief, while an unobserved change leaves the prior belief intact. For goal and action questions, UserHarness applies $\pi(G^u,B_t^u)$: if the user believes a candidate satisfies the goal, the user proceeds with that candidate; if the user believes it fails, the user searches, avoids, repairs, or chooses a different action. This instantiation relates to the causal and precondition side of \Cref{fig:method}: a visible result can update belief when the story explicitly links the result to a causal consequence, but unrelated distractors do not.

In our main setting, the tested agent helps identify the queried user, target state, observed or unobserved update, and the option claim corresponding to each possible belief/action/goal. The strict evaluation disables option-provenance shortcuts, so the agent and harness must ground the answer in the story-level transition rather than in metadata about which option was generated as ``prior'' or ``updated.''

\subsection{Hi-ToM}

Hi-ToM stresses the right side of the belief box in Figure~\ref{fig:method}: nested belief. The environment $E_t$ contains rooms, users, object locations, entrances and exits, object movements, and public or private statements. The observation $O_t^u$ is determined by room co-presence and communication access. The belief state $B_t^u$ may contain high-order paths such as $B_t^{u\rightarrow v\rightarrow w}$, depending on the question order. The benchmark mostly queries belief rather than action, so $G$ and $A_t$ are used primarily to interpret story events that update the environment, while the final answer is a query over nested belief.

The key belief update rule applied here is visibility-preserving propagation. If two users co-observe an object movement, both update their own beliefs, and each can update beliefs about what the other observed. If a user leaves before a movement, that user's own belief remains at the last observed location, and other users who witnessed the departure can reason that the absent user did not see later changes. Public claims can update all present listeners; private claims update only the recipient. Thus, each nested belief path is updated only by events visible along that path.

The tested agent here is more as a branch-cutting translator. It identifies the queried belief chain, the target object, the relevant entrance/exit and movement events, and which users could know which events were observed. The auditor then checks whether the proposed nested belief requires impossible access to an event. Through our observations, the remaining errors in the main results are mostly concentrated in higher-order Hi-ToM questions, which is consistent with this benchmark being the deepest instantiation of nested $B_t$ rather than a simple own-belief loop.

\subsection{ToMi}

ToMi is a generated false-belief environment with people, rooms, containers, objects, and object transfers. The environment $E_t$ is the true location of each object and the presence state of each user. The observation $O_t^u$ records which object movements the user sees while present. The belief $B_t^u$ is the user's belief about object location, along with memory of earlier locations when the question asks for memory. The goal $G^u$ is usually search or retrieval, and the action $A_t^u$ is the container or room the user will search according to belief.

This benchmark directly relates to the classic false-belief part of \Cref{fig:method}. If the user sees an object move, $B_t^u$ updates to the new location. If the user leaves and the object moves afterward, $B_t^u$ preserves the old location even though $E_t$ has changed. Reality questions query $E_t$; memory questions query an earlier belief or observed state; belief questions query $B_t^u$; and search questions query $\pi(G^u,B_t^u)$.

Nested belief is not the dominant component for ToMi in our instantiation. The main use of the tested agent is therefore not to solve complex social recursion, but to translate the story into the correct target user, object, and last observed location, and then audit that the answer is from belief rather than reality. Our method can solve ToMi exactly because UserHarness cleanly exposes the observation--belief distinction.

\subsection{MMToM-QA}

MMToM-QA instantiates the loop as household task inference. The environment $E_t$ contains rooms, containers, appliances, objects, and task-relevant preconditions such as whether an item is available, usable, clean, open, repaired, or in the expected location. The observation $O_t^u$ is built from the user's action trajectory: which places the user inspects, what visible results they obtain, and which preconditions become resolved or blocked. The belief $B_t^u$ is the user's belief about where useful objects are and whether the task can proceed. Nested belief is generally not required. The goal $G^u$ is a household task, and $A_t^u$ is the next action that follows from the user's belief and goal.

The important update rule is action-consistency. Searching a container without stopping can rule out the target being there from the user's perspective. Observing that a precondition is fixed, available, or satisfied updates $B_t^u$ and licenses proceeding with the task. Conversely, if the user still believes the current candidate fails the goal, continued search or repair is expected. The final answer is therefore not merely an object-location query; it is a consistency check between belief, goal, and action.

The tested agent helps translate long household trajectories into the small set of relevant events: the likely goal, the target object or precondition, which searches succeeded or failed, and which action option matches the belief-driven next step. The auditor checks whether each option contradicts the inferred action rule $\pi(G^u,B_t^u)$. This benchmark directly relates to the step 3 and 4 in \Cref{fig:method}: goal and belief jointly determine action, and action feedback updates the environment for the next step.

\subsection{MuMA-ToM}

MuMA-ToM emphasizes communication, social intention, and belief of goal. The environment $E_t$ contains users, requested objects, object locations, utterances, and later outcomes. The observation $O_t^u$ includes messages heard by a listener and visible success or failure after the listener acts on the message. The belief $B_t^u$ includes the listener's own belief about the object location and, for belief-of-goal questions, what one user believes another user wants. The goal $G^u$ may be a listener's goal of locating an object or a speaker's social goal of helping or hindering. The action $A_t^u$ may be an utterance, a search action, or a later behavior revealing whether the communication was useful.

The communication update rule is central in this benchmark. A statement changes the listener's belief only if it is heard. The social interpretation of the statement is then checked against the environment and the resulting action: a message that guides the listener toward the correct object location supports a helping intention; a message that sends the listener away from the correct location supports hindering or prevention; and least-likely questions invert this consistency relation. Belief-of-goal questions use nested belief, but usually at a shallow level: what the speaker believes about the listener's goal.

The tested agent helps extract the speaker, listener, requested object, stated location, true or observed location, and the downstream outcome. The auditor then checks whether the proposed social intention is compatible with the communication event and the listener's belief update. In terms of the figure, MuMA-ToM activates both the observation-to-belief arrow and the belief/goal-to-action arrow: the utterance enters as an observation for one user, but it is also an action revealing another user's social goal.

\section{Experiment Details}
\label{apdx:exp_details}

\subsection{Experiment Setting Details}
\label{apdx:exp_settings}

All pure prompting baselines use the same basic framework: the tested model receives the story/question text and the answer options, and it must end with a single option letter. For these baselines, symbolic traces, prover outputs, and option-provenance metadata are hidden. The final-answer parser accepts only the requested option format; unparsable responses are counted as failures in the reported full-set numbers.

\paragraph{Direct prompt.}
The Direct Agent baseline is the minimal prompting condition. It asks the tested agent to answer from the given story without exposing the UserHarness variables or any symbolic transition trace. This baseline measures how well the model solves each ToM example using its own reasoning. Because the instruction does not force a distinction between the true environment and the queried user's belief, it is vulnerable to perspective leakage: the model may answer from the full narrative state rather than from what the user observed or believed. In the experiments, this prompt is used with deterministic decoding and with exactly the same answer options as the other prompt-only baselines.

\begin{tcolorbox}[
  enhanced,
  breakable,
  width=0.98\linewidth,
  colback=white,
  colframe=purple,
  boxrule=1.2pt,
  arc=6pt,
  left=5pt,right=5pt,top=4pt,bottom=2pt,
  title={\small Direct Agent prompt template},
  coltitle=white,
  colbacktitle=purple,
  fonttitle=\bfseries,
]
\small
\begin{lstlisting}[style=jsonTiny]
System:
Answer the question using only the given story.

User:
{question}
{options}

End your reasoning with exactly one line: Final answer: <one of {allowed}>
\end{lstlisting}
\end{tcolorbox}

\paragraph{Paradox prompt.}
The Paradox Prompt keeps the model as the sole solver but changes the decision procedure from direct answering to option-wise contradiction checking. For each candidate answer, the model is instructed to assume that the option is true and test whether this assumption conflicts with the story's environment, observations, beliefs, goals, communications, or actions. This prompt is meant to reduce obvious inconsistencies, especially when an option would require a user to know an unobserved event. However, it still has no persistent transition ledger and no executable belief state, so the model must remember and update all relevant states in free-form text.

\begin{tcolorbox}[
  enhanced,
  breakable,
  width=0.98\linewidth,
  colback=white,
  colframe=purple,
  boxrule=1.2pt,
  arc=6pt,
  left=5pt,right=5pt,top=4pt,bottom=2pt,
  title={\small Paradox Prompt template},
  coltitle=white,
  colbacktitle=purple,
  fonttitle=\bfseries,
]
\small
\begin{lstlisting}[style=jsonTiny]
System:
Solve theory-of-mind questions by option-wise contradiction checking. For each option, assume the option is true and test whether it contradicts the story's environment, observations, beliefs, goals, communications, or actions.

User:
{question}
{options}

Analyze each option one by one. Choose the option with no contradiction. End with exactly one line: Final answer: <one of {allowed}>
\end{lstlisting}
\end{tcolorbox}

\paragraph{Ledger prompt.}
The Ledger Prompt is the strongest pure prompting baseline. It gives the tested agent the same conceptual vocabulary used by UserHarness, including environment state, observations, beliefs, goals, actions, communications, and transition rules, but the ledger exists only in the model's generated reasoning. The model is instructed to derive the answer from a state-transition timeline, identify the prior state and belief update, and reject options that contradict this inferred ledger. This isolates the value of an executable harness: if natural-language ledger prompting were sufficient, this baseline would approach UserHarness. In practice, the ledger prompt improves some local reasoning but can still lose nested beliefs, invert option letters, or treat salient true-state facts as if they were available to the queried user.

\begin{tcolorbox}[
  enhanced,
  breakable,
  width=0.98\linewidth,
  colback=white,
  colframe=purple,
  boxrule=1.2pt,
  arc=6pt,
  left=5pt,right=5pt,top=4pt,bottom=2pt,
  title={\small Ledger Prompt template},
  coltitle=white,
  colbacktitle=purple,
  fonttitle=\bfseries,
]
\small
\begin{lstlisting}[style=jsonTiny]
System:
You are the model driver for a unified theory-of-mind prover. Track environment state, observations, beliefs, goals, actions, and communications. Apply these transition rules strictly:
1. direct observation updates an agent's belief to the observed resulting state;
2. lack of observation preserves the prior belief;
3. agents act from their current beliefs and goals, not hidden truth;
4. if an unchanged belief identifies a specific candidate as satisfying the goal, the agent exploits that candidate (collects, uses, studies, grabs, proceeds with it) instead of searching again;
5. if an observed change resolves a goal precondition, such as an item becoming available, safe, dry, usable, open, or fixed, the agent proceeds under that changed state instead of preserving the old blocked/waiting action;
6. visible causes or visible effects license the stated causal consequence when the story explicitly links them;
7. unrelated distractor events do not change target beliefs. Use the story only.

User:
{question}
{options}

Derive the answer from the state transition timeline. First identify the prior state, the environment change, whether the target agent observed the change or its visible result, the updated belief, and the action/goal implied by that belief. Then check each option by assuming it is true and rejecting any option that contradicts the transition ledger. If the target observed the visible result of a change, treat the causal consequence stated in the story as known enough for the belief update. If the agent still believes the current candidate satisfies the goal, choose the exploit/proceed action that follows from that belief and goal; treat continued search as contradictory unless the agent believes the candidate fails the goal. If the observed change resolves an obstacle or precondition, choose the action that proceeds with the task under the new state; cleanup or repair caused by the observed event can be part of proceeding.

Before the final line, explicitly match your derived belief/action/goal to the option text; do not invert the letter after deriving the correct state. End with exactly one line: Final answer: <one of {allowed}>
\end{lstlisting}
\end{tcolorbox}

\paragraph{Other Baseline Methods.}
In addition to the baselines described above, we report results for existing methods following the evaluation setting in AutoToM~\citep{zhang2026autotom}, which evaluates on the same five ToM benchmarks. We omit several older baselines whose reported performance is substantially below the methods considered here, in order to focus the comparison on the strongest and most relevant prior approaches.

One important distinction is that AutoToM uses multimodal inputs for some benchmarks, combining visual information with textual descriptions when available. In contrast, our method reconstructs the user's mental state using only textual information. Therefore, our method operates with strictly less input information on those tasks. Despite this disadvantage, UserHarness still achieves stronger performance than the reported baselines, including AutoToM. All model outputs are generated deterministically. Following the original AutoToM setting, the core model used by AutoToM is GPT-4o; we do not additionally evaluate AutoToM with stronger backbone models. Notably, even with GPT-4o as its backbone, AutoToM is outperformed by UserHarness using Qwen3-8B, further demonstrating the effectiveness of our framework.

\subsection{Data Statistics}
We evaluate UserHarness on the union of five widely used ToM benchmarks, normalized into a shared text-only multiple-choice schema for a total of 15{,}900 examples: 12{,}000 from ToMi (3{,}000 each of belief, memory, reality, and search questions), 1{,}200 from Hi-ToM (240 examples per belief order $0$ through $4$), 1{,}200 from BigToM (200 examples per question type $\times$ observed/hidden visibility cell), 900 from MuMA-ToM (202 belief, 496 belief-of-goal, and 202 social-goal items), and 600 from MMToM-QA (300 belief queries and 300 goal queries spread over the type 1.1--2.4). Each example carries a story, a question, the original answer options, and a metadata record that exposes the diagnostic slicing variables used in our analysis without leaking the gold answer.

All five benchmarks are publicly released research artifacts intended for academic evaluation of language-model Theory-of-Mind reasoning, and we use them strictly in this intended research context: we run them only as read-only test sets for inference-time evaluation, do not train on them, and do not redistribute the source data. The benchmarks contain synthetically generated stories about fictional first-name characters in household or simple social scenarios; they do not contain real-person identifiers, contact information, or offensive content. We did not collect any new human data, and our pipeline does not transmit benchmark text to external services beyond the model APIs already required for inference, so no additional anonymization step was needed.

\section{Analysis Details}

\subsection{Symbolic Harness vs.\ Model Harness: Setting Details}
\label{apdx:symbolic_ablation}

This section documents the experimental setup behind the Symbolic-vs-Model UserHarness comparison reported in \Cref{tab:symbolic_gap}. The aim of the ablation is to disentangle two sources of UserHarness's gain: the structure of the harness itself (the perception--belief--action loop, the per-benchmark instantiation rules, and the option-level proof procedure) from the residual contribution of the tested model when the rule scaffold cannot decide.

\paragraph{Pure-symbolic pipeline.}
The pure-symbolic harness reuses the same benchmark tasks as the main UserHarness setting and follows the loop in \Cref{fig:method}, but it never issues a model call. Story parsing, agent and event extraction, observation filtering, belief and nested-belief updates, action-consistency checks, and option-level proof are all carried out by deterministic procedures. Each benchmark plugs in the per-benchmark instantiation rules described in \Cref{apdx:benchmark_instantiations}: BigToM causal-and-precondition checks, Hi-ToM nested-belief propagation under co-presence and public/private claims, ToMi false-belief tracking by entrance/exit and movement events, MMToM-QA action-consistency search elimination, and MuMA-ToM communication-scoped belief updates with helping/hindering interpretation. The pipeline is calibrated only against the visible inputs of the test sets: rule heuristics inspect the surface form of the questions, options, and stories to refine grammar coverage, but they never read the gold labels. Any improvement is therefore attributable to scaffold design rather than to label leakage.

\paragraph{Process of bstention.}
When the rule procedure cannot decide an example, for instance, a Hi-ToM question whose nested chain includes an unrecoverable visibility cutoff, or a MuMA-ToM social-goal question whose communication content cannot be extracted from the surface text, the symbolic pipeline always emits a controlled abstention. For scoring purposes, the abstention is mapped to the most consistent default option, so the benchmark numbers in \Cref{tab:symbolic_gap} reflect a single deterministic answer per example rather than a partial coverage rate.

\paragraph{Comparison anchor.}
Because the model-participating UserHarness scores in \Cref{tab:main} cluster within a 3.65 point band, we anchor the gap analysis to a single backbone (Qwen3-32B) rather than tabulating every model. Qwen3-32B is the strongest open-source backbone in our setup with a complete UserHarness main run, and its per-benchmark accuracies sit close to the macro-best Claude-Opus-4.7 row, so the qualitative gaps are stable when the anchor is swapped for any of the other model-participating runs.

\paragraph{Gap computation.}
For each benchmark $b$, the gap reported in \Cref{tab:symbolic_gap} is computed as $\Delta_b = a_b^{\text{model}} - a_b^{\text{sym}}$, where $a_b^{\text{model}}$ is the Qwen3-32B UserHarness accuracy and $a_b^{\text{sym}}$ is the pure-symbolic accuracy. The macro gap is the mean of the per-benchmark gaps, not the difference of the macro accuracies.

\subsection{Audit Override Calibration: Setting Details}
\label{apdx:override_ablation}

This section documents the audit-override ablation behind \Cref{fig:calibration}. The ablation isolates whether models can reliably overrule the harness when they disagree with its proof, which is the natural failure mode of giving the tested agent unrestricted authority over the final answer.

\paragraph{Two paired conditions.}
We report two paired runs per backbone. The control condition keeps the main UserHarness audit signal but does not let the model alter the proof: it simply records whether the audit accepted, rejected, or abstained. The override condition is identical except that an audit decision of \texttt{REJECT} is allowed to replace the harness answer with the model's preferred option. All other settings, including translation prompt, proof prompt, specialist branches, decoding temperature, and normalized inputs, are held fixed across the two conditions, so any macro-accuracy change is attributable to the override step alone.

\paragraph{Calibration statistics.}
Two complementary quantities drive the right panel of \Cref{fig:calibration}. (i) \emph{Rejected-proof correctness} is the fraction of rejected harness proofs that were nevertheless correct under the gold answer; it measures how often the audit rejects a trustworthy UserHarness proof. (ii) \emph{Override precision} is the fraction of accepted overrides that produced the correct answer; it measures how often a free-form override actually corrects a real mistake. A well-calibrated audit should keep the first quantity low and the second high, but in our experiments every backbone shows the opposite ordering. The macro-accuracy delta annotated in the left panel of \Cref{fig:calibration} is the difference of macro accuracies between the override and control conditions.

\subsection{Effective Compute: Setting Details}
\label{apdx:effective_compute}

This section documents the compute accounting behind \Cref{fig:compute}, which compares accuracy against the model-output budget that each method actually spends on the final answer.

\paragraph{Effective vs.\ nominal compute.}
The horizontal axis of \Cref{fig:compute} is the number of effective model-output tokens. Per-method rules are as follows. Prompt-only methods (Direct, Paradox, Ledger) count every prompt-and-response token, since the model owns every answer. The pure symbolic harness uses zero model-output tokens. UserHarness main setting counts translation, proof, specialist, and tie-break tokens.

\paragraph{Token approximation.}
To make comparisons consistent across providers and tokenizers, all token counts are computed by the same regex tokenizer (\texttt{\textbackslash w+|[\textasciicircum\textbackslash w\textbackslash s]}) over the rendered prompt and response strings. This is a uniform proxy rather than a billed-token count; it does not match any single provider's tokenizer exactly, but it preserves the ordering of methods and is easily reproducible.

\subsection{Difficulty Tiers and Error Categories: Setting Details}
\label{apdx:difficulty_tiers}

This section documents the slice-level breakdown shown in \Cref{fig:difficulty_tiers} and the structural error categories that the main paper attaches to each panel. The aim is to reveal what kinds of ToM examples remain hard once the harness scaffold is in place.

\paragraph{Per-benchmark slicing variables.}
Each benchmark is sliced by a diagnostic variable that is available before scoring and that captures a meaningful structural property of the example: BigToM by question type (belief, goal, action) crossed with whether the queried update is observed or hidden; Hi-ToM by belief order, where order $0$ is a reality query and order $k$ ($k\geq 1$) is a depth-$k$ nested belief; MMToM-QA by canonical question type (1.1--1.3 are belief queries about object location given a partial action trajectory; 2.1--2.4 are goal queries about which item the user is searching for; MuMA-ToM by the benchmark's own label (belief, belief of goal, social goal); and ToMi by generated question type (belief, memory, reality, search). The slicing variables are read from the metadata fields of each test data record.

\paragraph{Tier definitions and figure choice.}
For interpretation, slices with accuracy above 98\% are labeled easy, slices in the $[90\%,98\%)$ range are labeled medium, and slices below 90\% are labeled hard. The main-paper heatmap in \Cref{fig:difficulty_tiers} restricts to the four non-saturated benchmarks (BigToM, Hi-ToM, MMToM-QA, MuMA-ToM) for legibility; ToMi is omitted because every slice is at $100\%$ for every backbone in our setup, and including a panel of identical green cells would consume horizontal space without adding information.

\paragraph{Error categories.}
The residual error patterns visible in the heatmap fall into a small number of structural categories, which we use as a vocabulary in the main text. (i) BigToM: \emph{model transition or option-grounding errors}, concentrated in observed-belief slices where the harness has to ground a verbal causal description into the option's belief target. (ii) Hi-ToM: \emph{nested-belief propagation errors} at orders $\geq 2$, where the visibility cutoff between consecutive observers is fragile. (iii) MMToM-QA: \emph{action and precondition search inference errors} that dominate type 2.4, where the goal cannot be uniquely identified from the user's trajectory alone. (iv) MuMA-ToM -- \emph{belief-of-goal} and \emph{social-goal errors}, which require composing the listener's belief update with the speaker's helping-versus-hindering intention.

\section{Use of LLMs}
In this work, LLMs are used strictly for research support rather than as sources of substantive content. Their use falls into: (i) serving as the models tested under UserHarness settings, and (ii) assisting with language refinement during paper writing. For writing support, we used ChatGPT solely to polish text (improving coherence and grammar) while all ideas, logic, results, and technical contributions originate from the authors. To safeguard rigor, we have carefully reviewed all LLM-refined texts to confirm that no hallucinated content was introduced and that the original arguments, findings, and perspectives were faithfully preserved.

\section{Clarification, Scope, and Significance}
\label{apdx:clarification_scope_significance}

\subsection{Why UserHarness Matters}

\paragraph{A Missing Structure in Theory-of-Mind Evaluation.}
A central challenge in Theory-of-Mind (ToM) reasoning is not merely producing a plausible answer, but recovering the mental position from which a user or agent acts. Many ToM failures arise because the model is given the full story and therefore has access to facts that the target user did not observe. UserHarness addresses this structural mismatch directly. Instead of asking a model to answer from the raw narrative, it requires the model to reconstruct what the target user observed, believed, intended, and could reasonably do. This makes the user's subjective state, rather than the omniscient story state, the primary object of inference.

\paragraph{From Answer Prediction to User-Mind Reconstruction.}
The main contribution of UserHarness is a change in the form of inference. Standard prompting methods often encourage models to reason in free-form text, where the true environment, the user's belief, and nested beliefs can easily become conflated. UserHarness separates these quantities explicitly through an observation--belief--action trace. This separation is important because many ToM problems are designed precisely around cases where reality and belief diverge. By forcing answers to be justified through the user's accessible mental trajectory, UserHarness provides a more faithful and verifiable route to ToM reasoning.

\paragraph{Why This Is More Than Stronger Prompting.}
UserHarness is not intended as another prompt template. Prompt-only baselines can ask the model to ``think from the user's perspective,'' but the model still internally controls the entire reasoning process and may leak privileged information from the full narrative. In contrast, UserHarness externalizes the reasoning structure: observations, belief updates, communication access, candidate actions, and option-level consistency are represented and checked explicitly. The empirical gains therefore come not from longer reasoning alone, but from constraining the model to reason through a protected user-perspective scaffold.

\paragraph{Why This Is More Than a Symbolic Shortcut.}
A natural concern is that explicit belief tracking may exploit regularities in existing ToM benchmarks. We view this concern as important, and our symbolic-only analysis is included precisely to make the issue measurable. The results show that structured belief tracking already solves some simpler or more regular benchmark regimes, which suggests that many ToM datasets contain recoverable logical structure. However, the large gap between symbolic-only and model-participated UserHarness on more complex settings also shows that symbolic rules alone are insufficient. Model participation remains necessary for semantic interpretation, ambiguous event translation, trace auditing, and cases where intentions or social goals cannot be reduced to simple object-location rules. Thus, UserHarness should be understood as a hybrid inference framework: symbolic structure reduces perspective leakage, while model participation handles linguistic and pragmatic complexity.

\paragraph{A Diagnostic Framework for Model and Benchmark Limitations.}
Beyond improving accuracy, UserHarness provides diagnostic value. It separates errors caused by missed observations, incorrect belief updates, weak nested-belief tracking, poor action prediction, and social-intention ambiguity. This decomposition makes failures more interpretable than a final-answer score alone. It also reveals when a benchmark is largely solvable by structured belief rules and when model-level semantic reasoning is still required. In this sense, UserHarness is useful not only as a method for improving ToM performance, but also as a tool for understanding what current ToM benchmarks actually measure.

\paragraph{Relevance to User-Centric Assistants.}
Although the experiments use controlled ToM benchmarks, the underlying problem is closely related to user-facing assistance. Real assistants often need to distinguish what is true from what the user knows, what the user may have missed, what the user intends, and how the user's future actions may follow from incomplete or mistaken beliefs. UserHarness studies this capability in a controlled setting where belief access and mental-state divergence can be evaluated objectively. We therefore view these benchmarks as a focused testbed for a broader user-understanding problem, not as a complete simulation of open-ended human interaction.

\subsection{Scope and Design Clarifications}

\paragraph{UserHarness Targets Controlled ToM Reasoning, Not All Human Social Understanding.}
Theory-of-Mind is broad, covering beliefs, desires, intentions, emotions, deception, norms, memory, and social pragmatics. UserHarness focuses on a narrower but fundamental part of this space: reconstructing a user's belief and intention state from accessible observations and using that reconstruction to answer ToM questions. This scope is intentional. By isolating observation, belief update, nested belief, and action consistency, UserHarness makes a difficult component of user understanding measurable and reproducible. It does not claim to solve every form of social reasoning or real-world user modeling.

\paragraph{Benchmark Instantiations Are Interface Specifications, Not Answer-Specific Tuning.}
Different ToM benchmarks present stories, questions, and answer options in different formats. UserHarness therefore requires benchmark-level instantiations that map each dataset's input structure into the common observation--belief--action trace. These instantiations should be understood as interface specifications: they define how events, users, observations, and candidate claims are represented. They do not provide answer labels to the system and are not intended to encode item-specific solutions. This design lets UserHarness remain general at the reasoning level while accommodating heterogeneous benchmark formats.

\paragraph{Why Model Participation Is Constrained.}
UserHarness deliberately limits the model's authority. The model helps interpret language, identify relevant events, and resolve incomplete traces, but the final answer must remain consistent with the reconstructed user-perspective state. This design choice is motivated by the audit results: models can be persuasive but poorly calibrated when freely overriding structured proofs. Constrained participation therefore balances the strengths of language models with the reliability of explicit state tracking. The goal is not to remove the model, but to route its reasoning through a structure that prevents common ToM failures.

\paragraph{Multiple-Choice Evaluation Is a Controlled Proof Setting.}
Many ToM benchmarks are formulated as multiple-choice tasks, and UserHarness uses the answer options as candidate claims to be checked against the reconstructed trace. This should be interpreted as a controlled proof setting rather than a claim that all user-mind reconstruction is naturally multiple-choice. The option-level formulation is useful because it allows objective comparison across models and baselines. At the same time, the trace itself is the central object: it records what the user observed, believed, and could infer before the answer is selected. Extending this framework to open-ended belief and action generation is a natural next step.

\paragraph{The Text-Only Setting Is an Intentional First Step.}
Some evaluated benchmarks are associated with multimodal or richer social settings. In this paper, we use text-only inputs to focus on the core mental-state reconstruction problem under a common language-agent interface. This choice avoids mixing ToM reasoning with separate challenges in visual grounding or multimodal perception. The results therefore should be interpreted as evidence for UserHarness in language-based ToM reasoning. Incorporating visual observations into the same user-mind trace is an important future direction.

\paragraph{High Accuracy Does Not Mean ToM Is Solved.}
The strong benchmark performance of UserHarness should not be read as a claim that current agents possess human-like Theory of Mind. Instead, the results show that a substantial portion of benchmark ToM reasoning becomes more reliable when the user's mental state is externalized and protected from privileged information leakage. The remaining errors, especially in higher-order belief, future-action prediction, and social-goal inference, indicate that richer representations are still needed. UserHarness therefore advances ToM evaluation by clarifying which parts can be stabilized through structure and which parts remain open challenges.

\paragraph{Compute Efficiency Comes From Structured Inference.}
UserHarness improves the accuracy--compute tradeoff because it reduces unnecessary free-form reasoning. Rather than asking the model to produce long explanations, the framework directs computation toward event translation, belief-state construction, and consistency checking. This does not mean that symbolic processing is free, nor that output tokens capture every possible cost. Rather, the key point is qualitative: more unconstrained language generation is not the main path to better ToM reasoning. Structured inference provides a more reliable way to spend computation.

\paragraph{Relationship to Existing ToM Methods.}
UserHarness is complementary to prior work on perspective prompting, perception-to-belief reasoning, temporal belief tracking, symbolic ToM, and hypothesis tracing. Its distinguishing emphasis is to treat user-mind reconstruction as the organizing object that links observations, beliefs, goals, actions, communication, and nested mental states. This broader trace-based formulation allows the framework to cover several ToM question types under a unified inference procedure, while also making the source of each answer explicit.

\paragraph{Final Takeaway.}
UserHarness is intentionally structured, controlled, and diagnostic. It does not claim that ToM can be solved by prompts alone, nor that symbolic rules alone capture the richness of human social reasoning. Its contribution is more precise: it shows that many ToM failures stem from a missing user-perspective state representation, and that explicitly reconstructing this state can substantially improve reliability across diverse benchmarks. By separating the true environment from the user's observations, beliefs, intentions, and nested beliefs, UserHarness provides both a practical inference-time scaffold and a clearer foundation for future user-centric assistants.

\end{document}